\documentclass{article}

\PassOptionsToPackage{numbers}{natbib}
\usepackage[preprint]{nips_2018}
\usepackage[utf8]{inputenc} % allow utf-8 input
\usepackage[T1]{fontenc}    % use 8-bit T1 fonts
\usepackage{hyperref}       % hyperlinks
\usepackage{url}            % simple URL typesetting
\usepackage{booktabs}       % professional-quality tables
\usepackage{amsfonts}       % blackboard math symbols
\usepackage{nicefrac}       % compact symbols for 1/2, etc.
\usepackage{microtype}      % microtypography
\usepackage[pdftex]{graphicx}
\usepackage{subcaption}

\newcommand{\cbrac}[1]{\left( #1 \right)}
\newcommand{\sbrac}[1]{\left[ #1 \right]}

\title{Deep Curiosity Loops in Social Environments}

\author{
  Jonatan Barkan\\
  Curiosity Lab, \\Department of Industrial Engineering\\
  Tel-Aviv University, Israel\\
  \texttt{jbarkan14@gmail.com} \\
  \And
  Goren Gordon \\
  Curiosity Lab, \\Department of Industrial Engineering\\
  Tel-Aviv University, Israel\\
  \texttt{goren@gorengordon.com} \\
}

\begin{document}

\maketitle

\begin{abstract}
Inspired by infants' intrinsic motivation to learn, which values informative sensory channels contingent on their immediate social environment, we developed a deep curiosity loop (DCL) architecture.
The DCL is composed of a learner, which attempts to learn a forward model of the agent's state-action transition, and a novel reinforcement-learning (RL) component, namely, an Action-Convolution Deep Q-Network, which uses the learner's prediction error as reward.
The environment for our agent is composed of visual social scenes, composed of sitcom video streams, thereby both the learner and the RL are constructed as deep convolutional neural networks.
The agent's learner learns to predict the zero-th order of the dynamics of visual scenes, resulting in intrinsic rewards proportional to changes within its social environment.
The sources of these socially informative changes within the sitcom are predominantly motions of faces and hands, leading to the unsupervised curiosity-based learning of social interaction features. The face and hand detection is represented by the value function and the social interaction optical-flow is represented by the policy.
Our results suggest that face and hand detection are emergent properties of curiosity-based learning embedded in social environments.
\end{abstract}

\section{Introduction}

Infants are masters of learning, as they assimilate vast amounts of novel stimuli through their active interaction with their environment.
Moreover, babies have an intrinsic motivation to learn and attend to the most informative channels available to them.
This curiosity drive is the hallmark of young children's behavior.
On the other hand, infants do not have full control over their surrounding, as locomotion comes much later in development.
Hence, the visual scenarios infants are embedded in is dictated by their caregivers, which in turn dominate the visual stimuli.
In other words, infants' visual scenarios are mostly composed of other social agents interacting with them \cite{fausey_faces_2016}.

The combination of the curiosity drive and visual social scenarios, found in infants, guided our novel Deep Curiosity Loop (DCL) architecture.
Artificial curiosity concerns with intrinsic motivation in the field of reinforcement learning (RL) wherein the reward is intrinsic to the agent, as opposed to extrinsically given by the experimenter \cite{schmidhuber_curious_1991,oudeyer_intrinsic_2007,oudeyer_intrinsic_2016}.
The reward represents some form of learning or learning progress, quantified by e.g. prediction error, reduction in prediction error, surprise, empowerment, etc. \cite{little_learning_2013}.
The curiosity loop is hence composed of a learner, which attempts to learn sensorimotor correlations, e.g. the forward model of the agent's interaction with its environment, and an RL component which receives the intrinsic reward, thereby resulting in a behavior that attempts to optimize the learning  process (of the learner) \cite{gordon_hierarchical_2012}.
In our Deep Curiosity Loop architecture, due to the visual and temporal stream of information, the learner is a deep convolutional neural network that attempts to learn the (actionable) dynamics of visual scenes.
The RL component is constructed as an action-convolution deep-Q network (AC-DQN), wherein an action is learned {\em for each pixel in the video stream}.

Mimicking infants' visual environment, we take as training data visual social scenarios, i.e. sitcom videos.
In these videos, the dominant activity is performed by social agents.
Combining these scenarios with the DCL architecture, we hypothesize that the agent will attempt to learn the dynamics of the video (learner) and hence will learn to attend, i.e. predict high value and select actions towards (AC-DQN), the most informative or surprising parts of the scene.
We hypothesize that faces are these most informative parts and confirm this hypothesis from our data, whereas hands are also a dominant social communication channel.
This results in the unsupervised (curiosity-based) learning of a value function that represents Socially-relevant Feature Detection (SFD), namely, face and hand detection, i.e. high value for faces and hands due to their informative nature, and a social interaction-optical flow, i.e. local actions converging to the social features in the scene as learned by the AC-DQN.

The main contributions of this paper are threefold:
(i) a novel deep curiosity loop architecture, that combines CNN with intrinsic reward and DQN;
(ii) a unique RL algorithm, i.e. action-convolution DQN that enables a highly parallelized RL scheme with ``passive'' video stream and;
(iii) an unsupervised learning of face and hand detection.

% The resultant face detection algorithm differs from conventional ones in several aspects.
% It is unsupervised, with no face labels at all; 
% it is an online algorithm, inspired by infants' continuous video stream input and;
% it is based on video as its source.

\section{Motivation and Related Works}

Previous studies have shown that human babies are able to see in the first few months of their life \cite{brown_development_1990} as well as to detect and recognize faces (\cite{jakobsen_efficient_2016}). 
First \citet{jayaraman_faces_2015} and later \citet{fausey_faces_2016} proved that in the first year of life the relevant environment has a distribution with disproportionately high probability to contain faces. 
They revealed this lopsidedness decreases as the year progresses and is replaced by another non-uniform distribution with a disproportionate amount of moving hands, often attached to objects, as it pushes into its second year.

Curiosity has been described as a predominant driver of human behavior and it had not gone unnoticed by the Artificial Intelligence community \cite{schmidhuber_curious_1991,schmidhuber_formal_2010}. 
The re-emerging field is highly related to developmental robotics, which is the interdisciplinary field that attempts to integrate developmental psychology, machine learning and robotics, by studying infant's learning behavior and implementing those insights in robots that learn by themselves \cite{schmidhuber_possibility_1990,kompella_continual_2015,oudeyer_intrinsic_2007,barto_intrinsically_2004,weng_developmental_2004,cangelosi_integration_2010}.  
This framework focuses on intrinsic motivation \cite{oudeyer_intrinsic_2007,barto_intrinsically_2004,singh_intrinsically_2010,kaplan_search_2007,simsek_intrinsic_2006}, where the reward of the agent does not come from an outside source but rather from internal processing. 
Thus, Artificial curiosity (AC) \cite{schmidhuber_possibility_1990,kompella_continual_2015,oudeyer_intrinsic_2007,barto_intrinsically_2004,weng_developmental_2004} was inspired by developmental psychology and attempted to create a curious robot, where curiosity is often defined as “behavior driven by learning” \cite{gordon_emergent_2014,gordon_curious_2012}. 
By rewarding the learning progress, which can be computed in several different ways \cite{little_learning_2013}, and using reinforcement learning (RL) algorithms \cite{sutton_reinforcement_1998} one can adapt the behavior so as to optimize the learning process.

One of the possible intrinsic reward functions is the history-agnostic prediction error of an internal model. 
\citet{gordon_hierarchical_2012} introduced the notion of Hierarchical Curiosity Loops (HCL), an architecture whereby each loop selects the optimal action that maximizes an agent’s learning of sensory-motor correlations. 
The intrinsic reward was produced by a learning object internal to the agent, namely a Learner, presented with the task of modeling the sensory-motor internal models, e.g. forward model, which means taking the current state and action of the agent and predicting the next state. 

It has been shown that this framework can create emerging exploratory behavior, such as the emergent appearance of proximo-distal maturation from purely intrinsic motivation \cite{stulp_emergent_2012}, in addition to complex arena-exploration motor primitives \cite{gordon_learning_2014}.  
Many interesting results have come from this field back into Neuroscience, attempting to explain specific developmental phenomena by observing them in robots \cite{gordon_curious_2012,stulp_emergent_2012,moulin-frier_curiosity-driven_2012}. Accordingly, the same framework has been suggested to give plausible explanation to animals' exploration, e.g. vibrissae movement and locomotive exploration in rodents \cite{gordon_emergent_2014,gordon_learning_2014}, as well as infants' behaviors, such as the order of phonetic learning \cite{moulin-frier_curiosity-driven_2012} and hand-eye coordinated movements \cite{gordon_curious_2012}.
While these examples differ slightly in their instantiation of the curiosity-based algorithm, they mostly differ in the environment the agent is embedded in.

In this work we wanted to simulate a similar environment to that of an infant in order to check our hypothesis that socially relevant features detection can be learned without preexisting knowledge and by using curiosity alone. 
In effect, we hypothesize the correctness of the equation:
\begin{eqnarray}
	\textnormal{Social Environment} \,\,+ &\textnormal{Curiosity}  = &\textnormal{Social Interaction Detector} \\\nonumber
    \textnormal{\small{(faces and hands=information)}} & \textnormal{\small{(information=value)}}&\textnormal{\small{(face \& hand detector)}}
\end{eqnarray}

% \citet{chentanez_intrinsically_2005} explored agents with intrinsic reward structures in order to learn generic options that can apply to a wide variety of tasks. In another paper, \citet{singh_intrinsically_2010} take an evolutionary perspective to optimize over the space of reward functions for the
% agent, leading to a notion of extrinsically and intrinsically motivated behavior. 
% \citet{schmidhuber_formal_2010} provides a coherent formulation of intrinsic motivation, which is measured by the improvements to a predictive world model made by the learning algorithm. Frank et al. [9] demonstrate the effectiveness of artificial
% curiosity using information gain maximization in a humanoid robot. Oudeyer et al. [20] categorize
% intrinsic motivation approaches into knowledge based methods, competence or goal based methods
% and morphological methods. Our work relates to competence based intrinsic motivation but other
% complementary methods can be combined in future work
The literature regarding social features detection is composed of two venues, namely, face detection and image segmentation.
Two main approaches to face detection currently dominate the field. The first is by feature selection following the seminal work of \citet{viola_rapid_2001} \cite{viola_detecting_2005}, the industry's benchmark due to its implantations in many popular libraries (such as OpenCV). Their face detection algorithm produces templates for faces by learning to select profitable features and training a classifier with them on a labeled dataset. It was the first to perform online inference rapidly. To its detriment, small differences in the same object necessitates specialized templates, most notably frontal and profile faces. This is due to susceptibility to non-symmetry (monotonicity) and extreme light conditions among others (\cite{li_statistical_2002}, \cite{froba_face_2004,yan_locally_2008} respectively).

The second approach is by way of Deep Learning methods, primarily Deep Convolutional Neural Networks (DCNN). 
In contrast with the previous framework, here the features are generated automatically and with respect to the data they are trained on. 
It has no a-priori knowledge about the features nor the areas it should prioritize to find patterns. Of these methods, the technique of using a region proposal network to suggest possible regions with faces is at the heart of many current state of the art methods, currently led by \citet{jiang_face_2017} who employ the Faster R-CNN \cite{ren_faster_2017} method. Both techniques lie in the realm of supervised learning, as they are provided with a set of positive and negative examples.

As for the unsupervised case, there has not been many successful attempts. \citet{le_building_2013} seminal work proved face detecting features, among others, can be learned with a Deep Autoencoder. Yet they did not arrive at a face detection model and had to measure each feature's performance in classifying faces. Furthermore, their network used 1 billion parameters and trained on 10 million images. A somewhat recent work by \citet{walther_unsupervised_2012}, aimed for the purpose of unsupervised face detection, has managed to perform well by leveraging Organic Computing. However, it has not been tested on real-world data and its processing time for a single image is close to 1 minute.

Regarding image segmentation, \citet{long2015fully} trained a fully convolutional network for image segmentation. While not focusing on socially relevant features exclusively, these were part of the labeled training set for the supervised learning procedure. The trained network and its adaptation were used in socially-specific scenarios in \citet{ben2017model} in order to detect social interaction in still images. However, all these attempts used fully labeled supervised learning algorithms.

% \citet{kompella_intrinsically_2017}
% \citet{kompella_continual_2017}
% \citet{kompella_autonomous_2012}
% \citet{eslami_attend_2016}
% \citet{gottlieb_information-seeking_2013}

% \citet{kulkarni_hierarchical_2016} use intrinsic rewards when the environment provides delayed rewards, as they adopt a strategy to first learn ways to achieve intrinsically generated goals, and subsequently learn an optimal policy to chain them together

\section{Model}

In this section we present our proposed architecture and its embedding environment.
The environment is a generator of RGB images that may come from a single robot's camera or another continuous video feed. 
Thus the images are successive in time. For this experiment, all images are of fixed height $h$ and width $w$ and have RGB channels. 
The state space $S$ is defined as the set of all possible images.
We constrain the action space $A$ to movement parallel to the axes and to a single step size. 
As a result, the possible actions we allow are staying in the same location (using the sign $\circ$) or moving $k$ pixels up ($\uparrow_k$), down ($\downarrow_k$) left ($\leftarrow_k$) or right ($\rightarrow_k$), with a fixed step size $k$.

\begin{figure}[ht]
\centering
\includegraphics[width=0.9\linewidth]{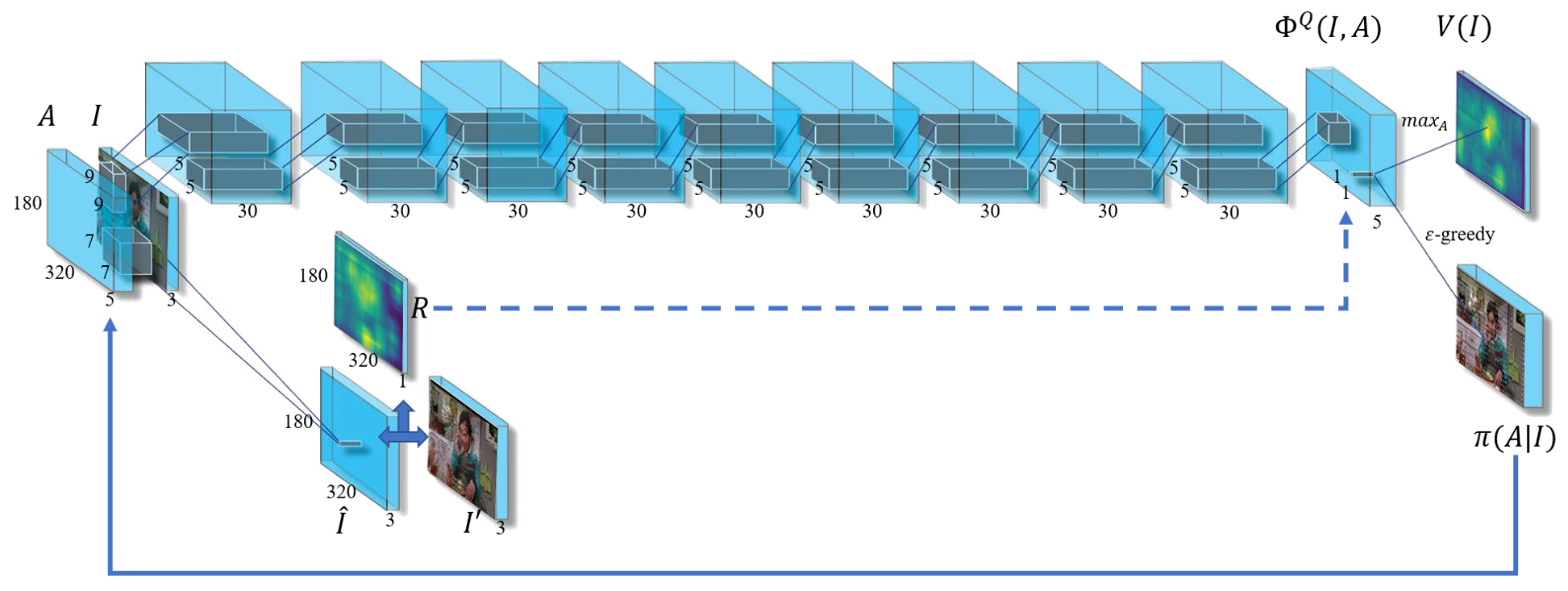}
\caption{The Deep Curiosity Loops architecture.}
\label{fig:faces change}
\end{figure}

In order to capitalize off of every single image, we introduce a novel approach which parallelize the agent's curiosity loop, surveying all potential states collectively. 
We do this with the simple trick of convolution, made possible by using Deep Convolutional Neural Networks (DCNN's) as the learning algorithms of the leaning elements of the Deep Curiosity Loop.
With this in mind we define the state space $\mathbb{S}$ as the set of all possible bounding boxes with predetermined $n$ and $m$ as height and width respectively, whereas the action space $\mathbb{A}$ becomes the set of matrices with entries from $\{ \circ, \uparrow_k, \downarrow_k, \leftarrow_k, \rightarrow_k \}$. 
We also add for notation $\hat{\textbf{A}}$ as the one-hot representation of an action matrix $\textbf{A}$ the new action space. 

% \begin{eqnarray}
% \label{NewStateSpace}
% S &=& \{ \cbrac{R, G, B} \mid R, G, B \in M_{h\times w} \cbrac{\sbrac{0, 1}} \} \\
% \label{StateSpace}
% \mathbb{S} &=& \{ \cbrac{R, G, B} \mid R, G, B \in M_{n\times m} \cbrac{\sbrac{0, 1}} \} \\
% \mathbb{A} &=& M_{h\times w} \cbrac{\{ \circ, \uparrow_k, \downarrow_k, \leftarrow_k, \rightarrow_k \} }
% \end{eqnarray}

% Immediately the Agent processes the available image, cropping a window with predetermined dimensions around its location. We call this window the "visible image" because its values form what the agent supposedly truly "sees". The centermost pixel indices of that window represent the agent's location. Let $U$ to be the agent's value space, containing all the possible visible images. Also let $B$ be the Agent's representation of the locations potentially reachable by it. For every $b \in B$ we define $U_t\cbrac{b}$ as the visual value of the Agent at time $t$ on location $b$. With regard to our primitive model, the location (equation \ref{LocationSpace}) and visual value spaces allow us to define a state as $s_t = \cbrac{b, U_t\cbrac{b}}$. Thus the state space is their Cartesian product (\ref{StateSpace}). The three spaces formal definitions are:
% \annote{start replace from here}

\subsection{Learner}

Our Learner $\Phi^L$ is a mapping that aims to learn a forward model of the agent's state-action transition. We chose a convolutional neural network with a single layer, without pooling and with a ReLU activation, that takes as input an image $I\in S$ and a one-hot encoding of an action, $\hat{\textbf{A}} \in \hat{\mathbb{A}}$ and outputs a predicted $\hat{I} \in S$, $\Phi^L(I, \hat{\textbf{A}})\mapsto\hat{I}$. Given the next image $I^\prime \in S$ the learner's cost function, $J^L$ is given by the Mean Squared Error between $\hat{I}$ and $I'$. 

\begin{equation}\label{LearnerLoss}
J_L \cbrac{I, \hat{\textbf{A}}, I^\prime} =  \frac{1}{3wh} \sum_{i, j, c} \cbrac{\Phi ^L \cbrac{I, \hat{\textbf{A}}}\sbrac{i, j, c} - I^\prime \sbrac{i, j, c}}^2
\end{equation}

It updates by Stochastic Gradient Descent with momentum \cite{botev_nesterovs_2016}.
Since this is a pure convolutional network, the output (image) is the same size as the input (image), wherein each pixel in the output is affected by $n\times m$ pixels in the input image, centered on it.

\subsection{Intrinsic Rewards Image}

In order to compute the reward we decompose its calculation to two parts. Instead of using the learner's loss directly, we devised a slightly different reward function. First we compute the mean squared difference with respect to each pixel's maximum color value. This results in an image-like value structure with the same dimensions as the input image. Second, we convolve the result with an averaging filter with the same dimensions and zero-padding such that it retains its size. This has the desired affect of averaging with respect to the receptive field of the AC-DQN (see below). This too results in an image-like 2d (no depth/color, or depth 1) structure where every pixel value represents the reward of the state that this pixel represents.

\begin{equation}\label{reward}
\textbf{R} \cbrac{I, \hat{\textbf{A}}, I^\prime} =  \frac{1}{wh} \sum_{i,j}  \cbrac{\max_c \Phi ^L \cbrac{I, \hat{\textbf{A}}}\sbrac{i, j, c} - \max_c I^\prime \sbrac{i, j, c}}^2
\end{equation}

% Although this deviates from the Curiosity loop, it is amenable to our implementation. To generate the correct rewards, instead of acting on the next frame before the Learner predicts we act on it afterwards while also acting on the prediction itself. Only then we compute the rewards for that time step. Despite looking like a divergence from the Curiosity Loop, this is only a detour. The reason is that for every kind of action this mimics its correct influence on the prediction. Say we want to predict the next frame given that the agent moved one way by $k$ steps, than the Learner should have generated the same prediction but with the entire image shifting the opposite direction by the same $k$. Put another way, the agent looking up by 5 pixels is equivalent to a static Learner's entire prediction shifting down by 5 pixels.

\subsection{Action-Convolution Deep-Q Network}

With the intention of parallelizing the Deep Curiosity Loop we devised a new algorithm called Action-Convolution Deep-Q Network (AC-DQN). 
It is a variant of Deep Q-Network (DQN) and Double Deep Q-Network (DDQN). These networks are essentially $Q$ function approximators that employ the Deep CNN structure and learn through gradient descent (ADAM \cite{kingma_adam:_2014}) to minimize the TD-error. As with both algorithms, we make use of Experience Replay (prioritize \cite{schaul_prioritized_2015}). Between layers we use instance normalization \cite{ulyanov_instance_2016} which does not require turning off after learning.

AC-DQN is a deep CNN architecture $\Phi^Q$ with an image as its input and a matrix with the same height $h$ and width $w$ as the reward. The network's last layer is of dimensions $h \times w \times \vert A \vert$, where $A$ is the action space.
We are using DDQN, hence the target $\textbf{T}$ is the matrix-addition of the reward matrix (``image'') and the discounted approximation of the output of $\Phi^Q$ on the next frame, and the TD-error is denoted as $\delta$.

\begin{equation}\label{TD}
\textbf{T} = \textbf{R} + \gamma \cdot \max_{\hat{\textbf{A}} \in \hat{\mathbb{A}}} \Phi^Q\cbrac{I^\prime, \hat{\textbf{A}}} \quad
\delta = \vert \Phi^Q \cbrac{I, \hat{\textbf{A}}} - T \vert
\end{equation}

The final AC-DQN version uses a DCNN with 10 convolutional layers. Each hidden layer has a filter depth of 30 with ReLU activations whereas the output layer has depth 5 to match the number of actions. This layer has a linear activation function. 
Each layer's filter dimensions is $5\times 5$ except the first for which is $9\times 9$. 
Effectively, the AC-DQN has a receptive field of $45\times 45$ with respect to the input layer. 

The novelty of our proposed AC-DQN architecture is in its output layer.
In traditional in DQN variants, given input $s$, the output of the network is a vector where the entry for each action $a$ represents the prediction of $Q\cbrac{s, a}$. 
However, in order to parallelize the Deep Curiosity Loop our AC-DQN's output is a matrix with the hight and width of the input image $I$ and with depth the size of the action space (here $\vert A \vert = 5$).
In essence, we compute the Q-value for {\em each pixel} (environment) in the input image. Consequently, AC-DQN's loss function is actually the mean of the TD-error over all pixels.

\section{Experiments}

\subsection{Data}
The original dataset consists of 10 seasons of the hit Television series The Big Bang Theory as our ``social visual scene'', most consisting of 24 episodes. Below we describe the preprocessing and the reasoning behind it. 
% It may not be a perfect match but it does contain a lot of faces. However it has large amounts of moving hands and other challenges such as the distance the activity is shot from. This is an issue because from close vantage points some parts can be predictably unpredictable while from others they might be mistaken for noise and vice versa, e.g. eyelids and human bodies. As a result, the atmosphere with which our agent comes in contact takes after the second year more then the first. Collectively, both will influence what is learned.

Contrary to traditional DL training, we attempt to mimic infants' environment and processing.
Hence, we use {\em online learning} and consider only actions {\em within} the image. 
The latter model eye movement as changing the center of the ``visible image'', whereas head shifts can be seen as corresponding to camera movement. For this reason we divided each episode into shots, each with a minimal movement of the camera. We have done so by upholding a threshold for the maximal amount of pixel turnover rate on the outskirts of the frames. We selected the threshold to be 25\%, and defined the area to be all pixels within distance 10 from at least one of the borders of the frame. We have chosen such a low threshold so to preserve many of the frames where a movement occurs on the edges while maintaining a dataset with little to no movement generated from a camera shift. For example, we maintained sequences in which people and objects enter and leave the frame. With this definition we moved to create the shots. 

Another decision we made was to reduce the frames per second displayed in the shots. According to \citet{carpenter_movements_1988}, saccade latency is around 200ms. Therefore a frame rate of 5 per second seems appropriate. Additionally, in order for a mini-batch to be decent sized, we artificially demanded each shot to retain a minimum of 20 frames and a maximum of 32. Meanwhile, we also resized the frames to a less demanding 320x180 RGB channel. It seemed to be the middle ground between low resolution while conserving enough details for the model to learn.

The last significant cleaning effort we made was to clear the data from artificial objects, such as running captions or subtitles. Therefore many of our episodes do not include the shots taken right after the opening credits. Finally, the shots were sorted chronologically by season and episode from which they are taken in order to feed the agent in the correct order of events. 

The final dataset consists of 203780 frames in 7855 shots of the television series The Big Bang Theory, each consisting of 20-32 frames of a video reduced to 5 frames per second 320x180 RGB channel. Furthermore we transformed the frames to 3 dimensional arrays of 32bits floating point numbers ranging from 0 to 1. 

\begin{table}
  \caption{Summary of dataset, 10 seasons of The Big Bang Theory.}
  \label{table:season_summary}
  \centering
  \begin{tabular}{cccccc}
  \toprule
  {} & \multicolumn{3}{c}{Number of Frames} &  {Shots} & {Episodes} \\
  \textbf{Season} &             mean &   std &  sum &  &  \\
  % \textbf{Season} &                  &       &        &       &         \\
  \midrule
  \textbf{1     } &            25.71 &  4.51 &  13.96K &   543 &      17 \\
  \textbf{2     } &            26.03 &  4.52 &  19.86K &   763 &      23 \\
  \textbf{3     } &            26.09 &  4.42 &  19.98K &   766 &      23 \\
  \textbf{4     } &            26.12 &  4.42 &  24.48K &   937 &      24 \\
  \textbf{5     } &            26.32 &  4.53 &  26.08K &   991 &      24 \\
  \textbf{6     } &            26.03 &  4.48 &  21.84K &   839 &      23 \\
  \textbf{7     } &            25.66 &  4.40 &  19.76K &   770 &      24 \\
  \textbf{8     } &            25.87 &  4.45 &  21.70K &   839 &      24 \\
  \cline{1-6}
  \textbf{9     } &            25.66 &  4.35 &  19.66K &   766 &      24 \\
  \textbf{10    } &            25.68 &  4.45 &  16.46K &   641 &      23 \\
  \bottomrule
  \textbf{Total } &   \textbf{25.94} &  \textbf{4.46} &  \textbf{203.78K} &   \textbf{7855} & \textbf{229} \\
  \\
  \end{tabular}
\end{table}

Table \ref{table:season_summary} shows the main parameters of the data. In this work, we split the data such that the training set contains the first 8 seasons and the validation and test set the last 2. It amounts to an 80/20 split with respect to seasons and a roughly the same (actually 82/18) with respect to the number of shots and number of total frames. 
% The distribution of frames for these sets is displayed in Figure \ref{fig:frames}.
As for the part of face detection, we split the Deep Curiosity Loop's test set in 2. The validation set now consists of 24 shots, 1 from each episode of season 9. The rest goes to the test set.
% To note, we did not use the popular 60/20/20 splits because we did not know in advance whether convergence can be reached with so few data points.

% In order to justify our choice of data, we need to show that it contains the prerequisite information channel. For us, that entails proving the existence of a face channel.
% As we stated in section \ref{VisnEnv}, by exploring the data manually we came to the conclusion that it mirrors a child's second year more so then its first few months, meaning that it has more moving hands with and without accompanying objects and also complete moving human upper bodies (see \ref{fig:subjects} for examples). That said, we show no proof because we did not find a dependable open sourced software that detects it as well as the one for faces.

\subsection{Faces are Informative}
Given this dataset, we can test our hypothesis that faces form the majority of the information channel, chosen to be represented by pixel-color difference.
For this purpose, we extracted the squared difference images between all directly successive frames. Meanwhile we used an out of the box face detector from OpenCV (which uses a variant of \citet{viola_rapid_2001}) to obtain the locations of the faces in each image. Next we computed the distribution of the differences within the proposed locations for faces and outside of it. As can be seen in Fig.~\ref{fig:face_detector}(inset), the mean squared change is significantly greater inside the region containing a face than outside of it ($Z=1.86\times 10^8$, $p<.0001$ Wilcoxon Signed-Ranks Test). 
This reflects our observation that this phenomenon was found to be correct in all but 3 pairs of successive frames.

This result shows that in general, larger rewards will come from pixels within the vicinity of a face than far from it. 
To put it differently, for a curious agent it is more interesting to look at a face than to look randomly anywhere else in the picture.
However, other socially relevant features are prominent in the dataset, e.g. hands.
Since we do not use labeled supervised learning, but curiosity-based learning, the resulting model cannot differentiate between different socially-relevant features and will mark faces as well as hands with high value, indiscriminately.

% training: Big Bang Theory
% test: FDDB
\subsection{Socially-relevant Feature Detection}

The simple learner we constructed $\Phi^L$ resulted in a straightforward convolution of a centered Gaussian (not shown).
In other words, the learner converged to predicting the zero-th order of the visual dynamics, namely, the predicted next image is (almost) identical to the current image.
This resulted in prediction errors that corresponded to movements or change within the video stream, as all static components are correctly predicted.
Thus the reward matrix $\textbf{R}$ effectively represents changes in the scene.
Taken together with the analysis of the dataset showing that faces are one of the major sources of change, we obtained a reward matrix corresponding to faces (and hands).
We wish to emphasize that this was {\em not pre-programmed} or known a-priori, but rather is the emerging property of a simple CNN forward-model learner and the statistics of the embedding environment.

The link between rewards and faces and hands allow us to derive from the AC-DQN part of the Deep Curiosity Loop a set of socially-relevant feature detection models. %, for which we give 2 measures of evaluations. 
We remind the reader that the Action-Convolution Deep-Q Network's output is $\Phi^Q\cbrac{I, A}$. However, contrary to previous variants of DQN, it is not a real-valued vector but a real-valued matrix with the width and hight of $I$ and depth the size of the action space (5 in this work).

\subsubsection{Face and Hands Mask}
% are successful models output a bounding box around the face. That is the true for the ground truth we used. But our model does not work that way. Granted, the nature of the averaging operator helps our model become rather continuous, yet it does not force the mask to be a box. Our mask is more fluid. This is exemplified in figure \ref{rewards:rewards}, when we remember that the Critic strives for the optimal value function which is presented there. That discrepancy might help our results but it might also hurt.

%Given a threshold value $t$ and an image $I$, w

Given an image $I$ (Fig.~\ref{fig:results}(a)), we define the value image as $V\cbrac{I} = \max_a \Phi^C\cbrac{I, \hat{\textbf{A}}}$, Fig.~\ref{fig:results}(b). 
We now show that, as a result of our chosen environment, $V$ can be used to form a face and hands detection model. To do so we generate a binary mask from it with respect to a given a threshold value $t$.

\begin{figure}[ht!]
  \centering
  \includegraphics[width=0.9\columnwidth]{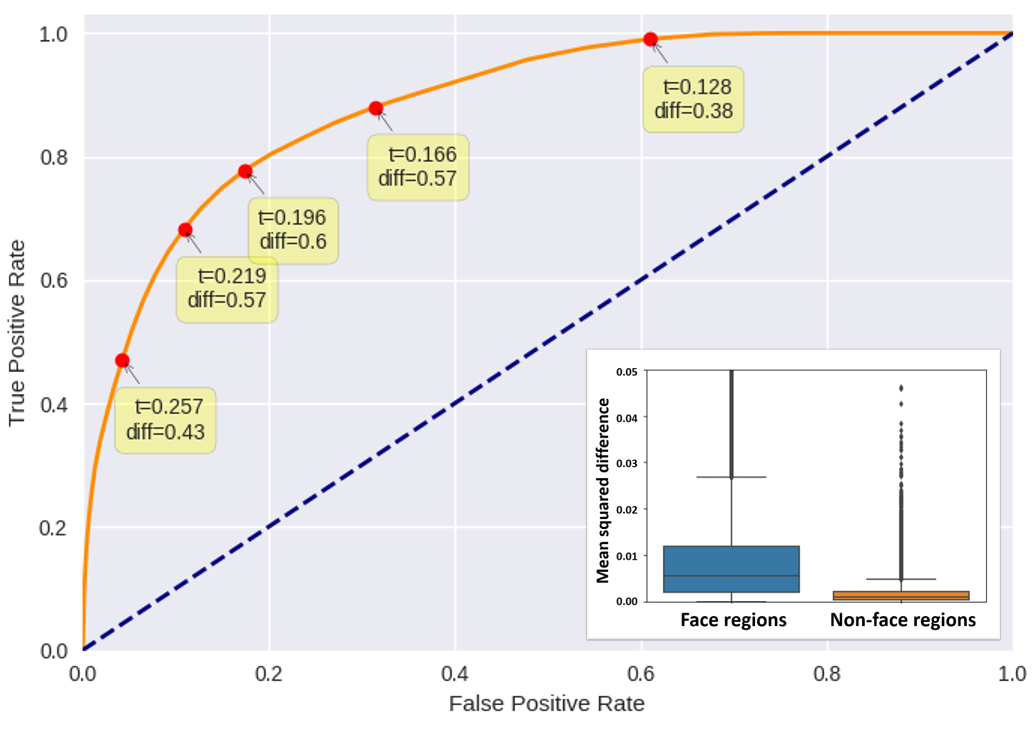}
  \caption{ROC curve of the value image, derived from the AC-DQN, with thresholds between -0.1 and 0.5.
  Inset: The mean squared difference between two successive frames, within a face region and outside of it. Cutoff at 0.05. A face region is detected using OpenCV frontal Viola-Jones face detector with 1.1 enlargement rate and with at least 6 neighbors. Performed on the first 6000 shots, only for frames where a face was detected, $\approx 100000$ pairs}
  \label{fig:face_detector}
\end{figure}

To test its worth as a face detector, we first picked a threshold by building a ROC curve (displayed in figure \ref{fig:face_detector}) from a validation set, where the ground truth is given by an OpenCV implementation of \citeauthor{viola_robust_2004}, the same algorithm with which we showed the existence of a face channel. We then measured the success of our detector by calculating Intersection Over Union (IOU) between it and our ground truth. The resulting AUC value of 0.877 shows that our unsupervised learned face detector works quite well.
Further analysis on the test set re-confirms this result.
However, it is important to note that since our model does not discriminate between faces and hands, a portion of the false negatives for face detection should still be considered as true positives for SFD.

\begin{table}[!ht]
\centering
\begin{tabular}{cccc}
\toprule
{Threshold} &       FPR &       TPR &		Youden index \\
\midrule
\textbf{0.128} &  0.595343 &  0.993316 &  0.397973 \\
\textbf{0.166} &  0.310977 &  0.951039 &  0.640061 \\
\textbf{0.196} &  0.179516 &  0.861723 &  0.682208 \\
\textbf{0.219} &  0.113833 &  0.753707 &  0.639873 \\
\bottomrule
\end{tabular}
\caption{Accuracy of four thresholds on the test set. We chose the Youden index as the optimality criteria, where it gives equal weight to the True Positive Rate (TPR) and the False Positive Rate (FPR) and is defined as $\max{TPR - FPR}$. We display the results of the $2:1$ ratios, i.e. $\max\cbrac{2\cdot TPR - FPR}$ and $\max\cbrac{TPR - 2\cdot FPR}$.}
\end{table}

% \begin{table}
% \centering
% \begin{tabular}{cccccc}
% \toprule
% {} & \multicolumn{3}{c}{Number of Frames} &  {} & {} \\
% \textbf{Season} &             mean &   std &  sum & Shots & Episodes \\
% % \textbf{Season} &                  &       &        &       &         \\
% \midrule
% \textbf{1     } &            25.71 &  4.51 &  13.96K &   543 &      17 \\
% \textbf{2     } &            26.03 &  4.52 &  19.86K &   763 &      23 \\
% \textbf{3     } &            26.09 &  4.42 &  19.98K &   766 &      23 \\
% \textbf{4     } &            26.12 &  4.42 &  24.48K &   937 &      24 \\
% \textbf{5     } &            26.32 &  4.53 &  26.08K &   991 &      24 \\
% \textbf{6     } &            26.03 &  4.48 &  21.84K &   839 &      23 \\
% \textbf{7     } &            25.66 &  4.40 &  19.76K &   770 &      24 \\
% \textbf{8     } &            25.87 &  4.45 &  21.70K &   839 &      24 \\
% \cline{1-6}
% \textbf{9     } &            25.66 &  4.35 &  19.66K &   766 &      24 \\
% \textbf{10    } &            25.68 &  4.45 &  16.46K &   641 &      23 \\
% \bottomrule
% \textbf{Total } &   \textbf{25.94} &  \textbf{4.46} &  \textbf{203.78K} &   \textbf{7855} & \textbf{229} \\
% \\
% \end{tabular}
% \caption{my caption}

% \end{table}
\label{table:accuracy_table}

\subsubsection{Bounding Boxes}

\begin{figure}
    \centering 
  \includegraphics[width=0.9\columnwidth]{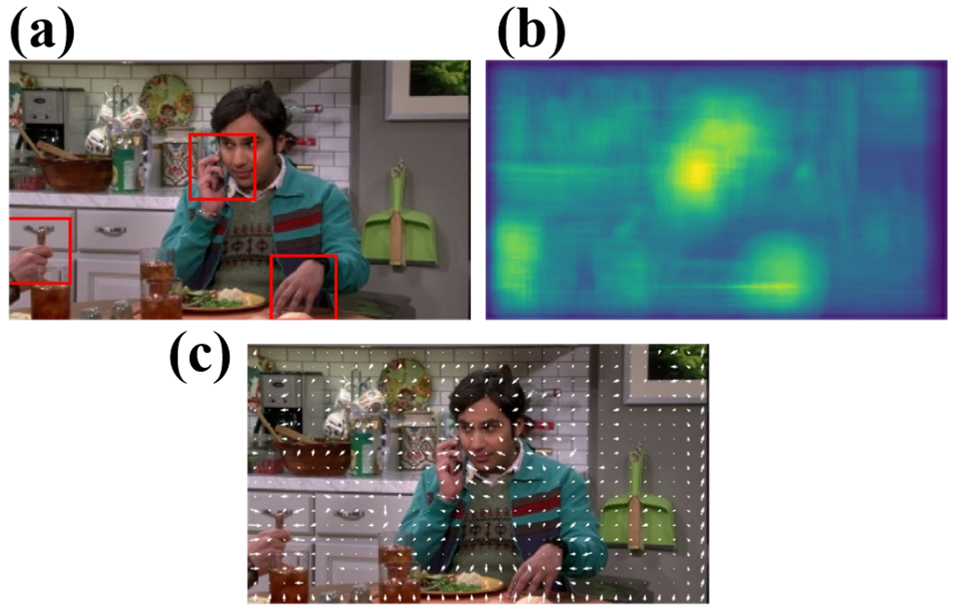}
\caption{Examples of the output of the DCL architecture.
(a) Input image with the resultant bounding boxes, $I$.
(b) Value function $V(I)$.
(c) Social Interaction Optical Flow, $\Phi^Q$.}
\label{fig:results}
\end{figure}

When it comes to face detection, most of the evaluation protocols rely on an output of a bounding box. Despite the fact that the DCL's output is not of this form, we have augmented it with the following processes. First we resize the image to our model's input dimensions ($180 \times 320$). Next we find all local maxima of $V\cbrac{I}$ that are not within 22 pixels from each other (22 because it is roughly half of the receptive field, which is 45 on each side) and produce a bounding box with each peak as the center of a $45 \times 45$ bounding box. Finally we translate those boxes to their location with respect to the original size of the image, Fig~\ref{fig:results}(a).

% and measure the performance on the widely used Face Detection Dataset and Benchmark (\citet{jain_fddb:_2010}) data set. In contrast to other face detection algorithms, not only is our model unsupervised but it did not train on any of the examples in the data set.

% The results are displayed ...

% \begin{figure}
% 	\centering
%     \includegraphics[width=\linewidth]{2002_09_06_big_img_11014}
%     \caption{Example of the two kinds of face detection models we get. On the left the bounding box (with 3 peaks), on the right the face mask. In the middle the image of the underlying $V$}
% \end{figure}

% \begin{figure}[!ht]
%     \centering 
% \begin{subfigure}{0.39\textwidth}
%   \includegraphics[width=\linewidth]{haar_y_curios_y}
%   \caption{haar y curious y}
%   \label{fig:1}
% \end{subfigure}\hfil
% \begin{subfigure}{0.39\textwidth}
%   \includegraphics[width=\linewidth]{haar_y_curios_y_2}
%   \caption{haar y curious y2}
%   \label{fig:2}
% \end{subfigure}

% \medskip
% \begin{subfigure}{0.39\textwidth}
%   \includegraphics[width=\linewidth]{haar_n_curios_y}
%   \caption{haar n curious y}
%   \label{fig:4}
% \end{subfigure}\hfil
% \begin{subfigure}{0.39\textwidth}
%   \includegraphics[width=\linewidth]{haar_y_curios_n}
%   \caption{haar y curious n}
%   \label{fig:5}
% \end{subfigure}

% \caption{Examples of success and failures of our model. }
% \label{fig:face_detection}
% \end{figure}

% figures: example (BBT), ROC (FDDB)
\subsection{Social Interaction-Optical Flow}

In addition to the SFD model, the policy $\pi(A \mid I)$ learned within the Deep Curiosity Loop provides a surrogate/closely related model. Learned via the novel Action-Convolution Deep-Q Network with the $\epsilon$-greedy algorithm, it amounts to a ``force field'' applied on the image. This is the result of the ``passive'' nature of the action, i.e. movement is performed within the image, together with a one-to-one correspondence between states and pixels. As can be seen in figure Fig.~\ref{fig:results}(c) the field is pointing towards the faces and the hands. We call this field the Social Interaction Optical Flow.

\section{Conclusions}
Attempting to mimic infant's learning processes and embedding environment, we developed a novel deep curiosity loop architecture. 
We have shown that faces are prominent motion sources within social scenes, hence, given a simplified forward model as a learner, results in rewarding faces.
However, since our proposed framework corresponds to curiosity-based intrinsic motivation learning, it is not supplied with labels and learns the value of all socially relevant features. 
These are not restricted to faces but also include hands and objects relating to them, comparable to children's normal visual development \cite{fausey_faces_2016}.

This lack of differentiation between faces and hands makes our model difficult to evaluate, since comparing to state-of-the-art face detectors \cite{jiang_face_2017} should not prove useful.
A more comprehensive comparison to a fully labeled and segmented dataset, as used by \citet{long2015fully}, could prove somewhat more adequate.

Our proposed framework is different from previous ones in several major aspects. 
Most notably, it attempts to learn {\em static} visual features from {\em dynamics} video feeds.
Furthermore, utilizing the curiosity loop, training a value function does not require an external reward function or labels.
Finally, while not mandatory, we have used online learning to better mimic infant's learning processes and still arrived at a valuable model representing socially-relevant features.

Future work will extend our training model in two dimensions. 
The first will include many more social scenes, e.g. Glee and Silicon Valley, in order to increase the generalization aspects of our model.
The second will replace normative social scenes with other more unique scenes, e.g. National Geographic, to explore which relevant features will emerge from the same framework.
Finally, deploying our proposed Deep Curiosity Loop architecture in a curious social robot might result in an infant-like learning process culminating in a socially engaging interaction.

\medskip

\small
   
\bibliographystyle{plainnat}
\bibliography{ArtificialCuriosity}

\end{document}